\documentclass[journal]{IEEEtran}
\usepackage{amsmath,amsfonts}
\usepackage{algorithmic}
\usepackage{algorithm}
\usepackage{array}
\usepackage[caption=false,font=normalsize,labelfont=sf,textfont=sf]{subfig}
\usepackage{textcomp}
\usepackage{stfloats}
\usepackage{url}
\usepackage{verbatim}
\usepackage{graphicx}
\usepackage{cite}
\usepackage{multirow}
\usepackage{booktabs}
\usepackage{adjustbox}
\usepackage[table]{xcolor}
\usepackage{pifont}
\usepackage{bbding}
\definecolor{Gray}{gray}{0.96}
\definecolor{revision_back}{HTML}{cbdbf6}
\definecolor{revision}{RGB}{0,0,255}

\begin{document}

\title{NumGrad-Pull: Numerical Gradient Guided Tri-plane Representation for Surface Reconstruction from Point Clouds}

\author{Ruikai Cui$^*$, Binzhu Xie$^*$, Shi Qiu\textsuperscript{\Envelope}, Jiawei Liu, Saeed Anwar, Nick Barnes
\thanks{R. Cui, J. Liu, N. Barnes are with the School of Computing, The Australian National University, Australia (e-mails: \{ruikai.cui, jiawei.liu3, nick.barnes\}@anu.edu.au). B. Xie, S. Qiu are with the Department of Computer Science and Engineering, The Chinese University of Hong Kong, Hong Kong SAR, China (e-mails: \{bzxie, shiqiu\}@cse.cuhk.edu.hk). S. Anwar is with the Department of Computer Science \& Software Engineering, the University of Western Australia, Australia (e-mail: saeed.anwar@uwa.edu.au).}
\thanks{$^*$ The authors contribute equally.}
\thanks{\textsuperscript{\Envelope}~Corresponding author: S. Qiu.}}

\markboth{Journal of \LaTeX\ Class Files,~Vol.~14, No.~8, August~2021}%
{Shell \MakeLowercase{\textit{et al.}}: A Sample Article Using IEEEtran.cls for IEEE Journals}


\maketitle

\begin{abstract}
Reconstructing continuous surfaces from unoriented and unordered 3D points is a fundamental challenge in computer vision and graphics. Recent advancements address this problem by training neural signed distance functions to pull 3D location queries to their closest points on a surface, following the predicted signed distances and the analytical gradients computed by the network. In this paper, we introduce NumGrad-Pull, leveraging the representation capability of tri-plane structures to accelerate the learning of signed distance functions and enhance the fidelity of local details in surface reconstruction. To further improve the training stability of grid-based tri-planes, we propose to exploit numerical gradients, replacing conventional analytical computations. Additionally, we present a progressive plane expansion strategy to facilitate faster signed distance function convergence and design a data sampling strategy to mitigate reconstruction artifacts. These components are synergistically integrated into a unified tri-plane-based pulling framework, in which numerical gradients, progressive expansion, and complementary sampling jointly address the locality and sparsity challenges of learning SDFs from unoriented point clouds. Our extensive experiments across a variety of benchmarks demonstrate the effectiveness and robustness of our approach. Codes are available at: \url{https://github.com/cuiruikai/numgrad-pull}.
\end{abstract}


\section{Introduction}
\IEEEPARstart{T}{he} signed distance function (SDF) has become a fundamental 3D representation for visual and spatial data modalities, widely adopted in various multimedia applications such as shape modeling, animation, virtual/augmented reality, \emph{etc.} Recent advances~\cite{neuralpull_ma,implicit_li,chen2023gridpull,ben2022digs} have increasingly focused on leveraging SDFs to reconstruct continuous surfaces from 3D point clouds. By mapping each 3D coordinate to a corresponding signed distance from the shape surface, SDFs implicitly define surfaces as zero-level sets, providing a powerful surface representation framework that offers distinct advantages in capturing high-fidelity 3D shapes with complex, arbitrary topologies.

Recent efforts~\cite{neuralpull_ma,ben2022digs,koneputugodage2024small,implicit_li,koneputugodage2023octree,pcp_ma,zhu2024ssp} in learning SDFs from point clouds tend to directly train neural networks on point sets, relying on extensive distance computations between query and surface points to guide the network towards accurate distance field convergence. A pioneering approach in this direction is Neural-Pull~\cite{neuralpull_ma}, which trains a neural network to iteratively project query points onto surfaces, using the learned signed distances and gradients to guide each time of adjustment. Subsequent works have extended this method by incorporating additional constraints for improved pulling target estimation~\cite{implicit_li,koneputugodage2024small,ben2022digs} or enhanced scalability of the approach~\cite{chen2023gridpull}. In general, existing methods often utilize deeper neural architectures and complex loss constraints to model signed distance fields, yet they still produce overly smoothed surfaces lacking fine-grained details and suffer from slow query speeds.

\begin{figure}
    \centering
    \includegraphics[width=0.82\linewidth]{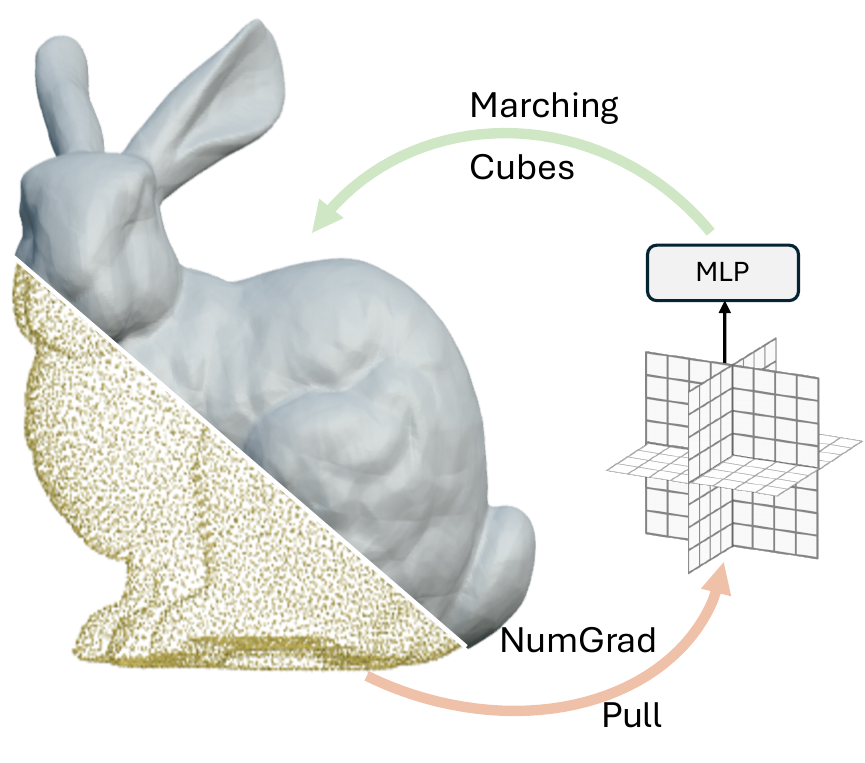}
    \caption{We present NumGrad-Pull, a tri-plane-based framework that enables efficient and robust surface reconstruction from unoriented point sets.}
    \label{fig:teaser}
\end{figure}

In this paper, we introduce NumGrad-Pull, a novel method for fast and high-fidelity surface reconstruction from point clouds. NumGrad-Pull models SDFs using a hybrid explicit–implicit representation, which combines a tri-plane structure that stores spatial information, with a shallow multi-layer perceptron (MLP) that maps feature embeddings on the tri-plane to a real signed distance value. Particularly, this design enables faster distance queries~\cite{chan2022efficient} compared to previous implicit methods~\cite{neuralpull_ma,implicit_li,koneputugodage2024small}, while maintaining strong geometric and shape expressive capabilities. Although the tri-plane structure improves reconstruction speed and fidelity, it poses challenges for stable training, since its grid-based format lacks feature interaction between adjacent grids, conflicting with the continuous nature of 3D surfaces. To improve the training stability of our approach, we innovate by using numerical gradients instead of analytical ones, allowing queries to back-propagate more effectively to relevant grid entities. Moreover, we introduce a progressive tri-plane expansion scheme, starting the training with a low-resolution tri-plane and gradually increasing the resolution. This approach accelerates convergence and helps prevent the network from getting into local minima.

Rather than using these designs as isolated modules, NumGrad-Pull integrates them into a unified tri-plane-based pulling framework for unoriented surface reconstruction. Within this framework, numerical gradients alleviate the locality of grid-based supervision, progressive expansion stabilizes coarse-to-fine tri-plane learning, and complementary sampling enhances the supervision of both near-surface details and global tri-plane regions. To assess the effectiveness of our proposed method, we evaluate it on widely used benchmarks that consist of both synthetic and real-world objects. Our main contributions are as follows:
\begin{itemize}
    \item We introduce a hybrid explicit–implicit tri-plane representation that significantly improves both speed and fidelity in learning signed distance functions from unoriented point clouds.
    \item We propose to utilize numerical gradients in place of traditional analytical gradients for back-propagating supervision on each query, significantly improving the stability and convergence of the signed distance function.
    \item We design a progressive tri-plane expansion scheme that further accelerates the convergence of tri-plane learning, enhancing both computational efficiency and reconstruction quality.
\end{itemize}

\section{Related Work}
Recent advancements in deep learning have highlighted the significant potential of neural networks for surface reconstruction from 3D point clouds. Below, we provide a brief review of methods relevant to point cloud-based surface reconstruction and surface representation.
\subsection{Traditional Surface Reconstruction}
Pioneering surface reconstruction methods like Delaunay triangulation~\cite{boissonnat1984geometric} introduce tetrahedral meshes to connect points, producing accurate surfaces for dense data but struggling with sparse or noisy inputs due to their computational intensity. Building on this, Poisson surface reconstruction~\cite{kazhdan2006poisson} formulates surface inference as a volumetric problem by solving the Poisson equation. This approach effectively produces smooth, watertight surfaces and handles noise well, but it often smooths out fine details. Moving least squares~\cite{levin1998approximation} is often used as pre-processing that smooths noisy data by fitting local planes; however, it also oversmooths high-frequency details. The interpolation-based approaches, such as Radial Basis Functions~\cite{carr2001reconstruction}, employ continuous functions to interpolate across sparse points, providing robust surfaces for smooth geometries, though computationally demanding for large datasets. By leveraging graph connectivity, graph-based methods address surface reconstruction by organizing point clouds through denoising, outlier removal, and robust normal estimation steps~\cite{amenta1998surface,carr2001reconstruction,huang2009consolidation}. These methods effectively handle sparse or non-uniform data, but can be computationally expensive, especially for high-resolution point clouds.

\begin{figure*}
    \centering
    \includegraphics[width=\textwidth]{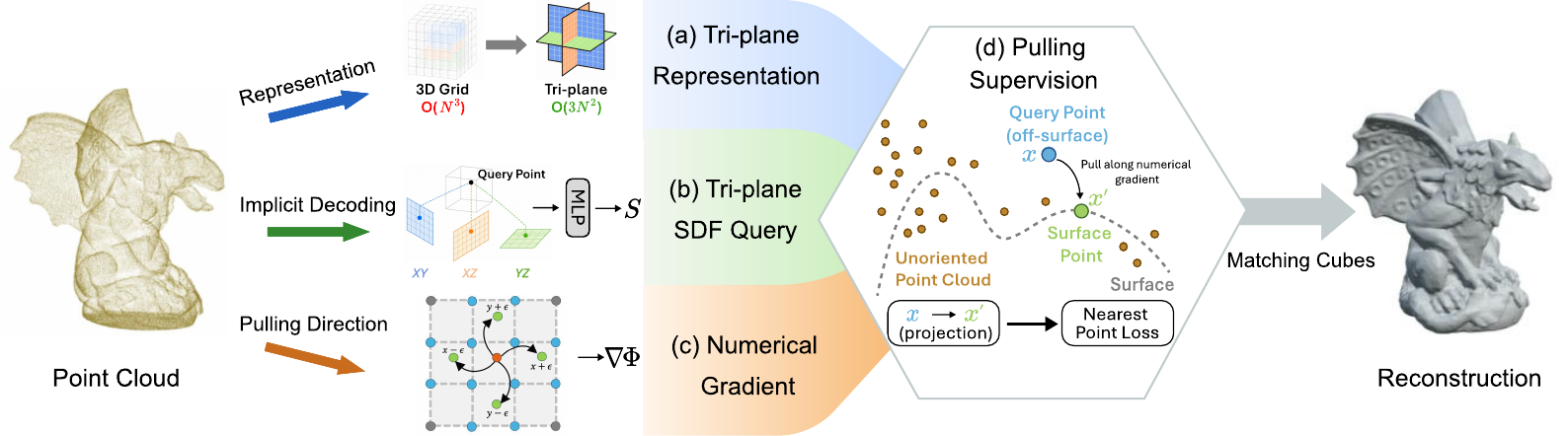}
    \caption{Overview of the NumGrad-Pull pipeline, which is built upon three synergistic components. (a) Unlike a dense 3D feature grid used in GridPull~\cite{chen2023gridpull} with $O(N^3)$ memory complexity, NumGrad-Pull factorizes spatial features into three axis-aligned 2D planes with $O(3N^2)$ memory. (b) For a given query point, our method extracts features from each of the three orthogonal planes via bilinear interpolation. (c) To address locality issues and stabilize training, we compute the gradient $\nabla\Phi$ numerically via finite differences, which involves adjacent grid entities in back-propagation and ensures smoother feature propagation across the tri-plane. Together, these three components are unified into a coherent pulling framework. (d) Using the signed distances and numerical gradients from the tri-plane-based SDF, the network is trained by pulling each off-surface query point $x$ toward its nearest neighbor $x'$ on the surface.}
    \label{fig:illustration}
\end{figure*}

\subsection{Neural Surface Reconstruction}
More recently, neural networks have enabled learning-based approaches to surface reconstruction, where implicit functions trained on large datasets facilitate detailed surface modeling even with sparse inputs~\cite{mescheder2019occupancy,park2019deepsdf,wang2025noisesdf2noisesdf}. However, estimating surfaces from unoriented point sets remains challenging without ground truth signed distance values.
Efficient neural field representations such as InstantNGP~\cite{muller2022instant} provide significant acceleration when signed-distance supervision is available, whereas normal-orientation methods such as GCNO~\cite{GCNO} can serve as a preprocessing step prior to training an oriented neural SDF. When applied to raw unoriented point clouds, however, these pipelines are inherently constrained by the quality of the proxy SDFs or estimated normals, leaving the final reconstruction vulnerable to preprocessing errors. 
Point2Surf~\cite{point2surf_erler} addresses this issue by learning a prior over local patches with detailed structures and coarse global information. Moreover, \cite{neuralpull_ma} introduce the first pulling operation-based surface reconstruction scheme, observing that the gradient of a neural SDF represents the normal direction of a query point. By using this gradient and the estimated distance from the SDF, any query point can be pulled to the surface, considering the neural SDF as a close approximation of the ground truth. Building on this scheme, Grid-Pull~\cite{chen2023gridpull} presents a discrete grid-based representation to improve the scalability of pull-based methods. Subsequent work by \cite{koneputugodage2024small} applies guiding points for incrementally changing optimizations toward the true surfaces. More recently, \cite{implicit_li} propose a bilateral filter to smooth the implicit field, retaining high-frequency geometric details. Different from existing studies, we are pioneering the exploration of hybrid explicit–implicit representations~\cite{chan2022efficient} for fundamental surface reconstruction problems, where our approach shows promise in accelerating query speeds and preserving geometric details across various reconstruction tasks~\cite{lin2023magic3d,cui2024lam3d, zhu2024pcf, zhu2025rethinking,qiu2025creating,xie2026egohandicl,hong2022pointcam,zhu2026cos3d}.

\subsection{Neural Surface Representation} 

Implicit representations have become widely used for surface reconstruction from point clouds, \emph{e.g.}, Neural-Pull~\cite{neuralpull_ma} devises a deep fully-connected network to represent the distance field as a continuous function. However, this approach results in slow query speeds, as each distance query requires a complete pass through the network. In contrast, explicit representations, such as discrete voxel grids~\cite{chen2023gridpull}, offer faster query times but face memory limitations, with memory consumption scaling as $O(n^3)$ as the discretization resolution increases. Given complementary advantages provided by explicit and implicit representations, motivated to develop a hybrid approach for surface reconstruction that combines the strengths of both paradigms. 
In this work, we present the first hybrid explicit-implicit method for neural surface reconstruction from point clouds, leveraging the tri-plane structure~\cite{chan2022efficient}, flexibility of implicit decoding~\cite{mildenhall2021nerf}, and the efficiency of explicit storage~\cite{muller2022instant,li2023neuralangelo}. Particularly, our proposed tri-plane method stores features on axis-aligned planes and deploys a lightweight implicit decoder to aggregate these features for efficient querying, achieving computational and memory efficiency. 
\section{Method}
\label{sec:method}

NumGrad-Pull reconstructs continuous surfaces from unoriented point sets by parameterizing a signed distance function (SDF) with a hybrid explicit-implicit tri-plane structure (Sec.~\ref{sec:method:triplane}). The network is trained with numerical gradients (Sec.~\ref{sec:method:numgrad}), which help stabilize the training process. Additionally, we introduce a progressive tri-plane expansion strategy (Sec.~\ref{sec:method:progressive}) and a novel query location sampling method (Sec.~\ref{sec:method:data}) to enhance both convergence speed and reconstruction quality. The overall pipeline of our proposed method is illustrated in Fig.~\ref{fig:illustration}. 

\subsection{Preliminaries}

A signed distance function is a function $\Phi: \mathbb{R}^3 \mapsto \mathbb{R}$ that maps a spatial position to a real value representing the signed distance to the shape surface. Using this function, the shape surface can be defined as the zero-level set of $\Phi$, allowing for surface extraction through methods such as Marching Cubes~\cite{lorensen1998marching}. To learn a neural SDF $\Phi$ from a given point cloud $P$ without normal information, \cite{neuralpull_ma} propose to query the function at various positions, and train a network to ``pull'' each query position $q$ toward the shape surface, approximated by the nearest point on $P$ to $q$. Specifically, this pulling operation is defined as:
\begin{equation}
\label{eq:pull}
    q' = q - \Phi(q) \times \frac{\nabla \Phi(q)}{\|\nabla \Phi(q)\|},
\end{equation}
where $\frac{\nabla \Phi(q)}{\|\nabla \Phi(q)\|}$ denotes the normalized gradient of $q$ and $|\Phi(q)|$ is its distance from the shape surface, both predicted by the SDF. In practice, neural SDFs are often implemented as deep fully-connected networks, which can be slow to query due to the need for a full forward pass regarding each time of a distance query. This limitation highlights the critical need for a faster SDF parameterization that achieves high reconstruction quality without compromising query efficiency.

\subsection{Tri-plane Formulation}
\label{sec:method:triplane}

Unlike prior methods~\cite{neuralpull_ma,implicit_li} that rely solely on deep fully-connected networks to learn a neural SDF, we leverage tri-plane embeddings to efficiently incorporate explicit 3D spatial information. Basically, our method estimates the signed distance of a query position by interpolating within the tri-plane structure and decoding through a shallow MLP to produce a real-valued signed distance output. By combining tri-plane structures with finite differentiation for gradient computation, our end-to-end trainable neural SDF model captures a high-fidelity 3D shape representation of a given point cloud.

To formulate our hybrid explicit-implicit representation for a neural SDF, we integrate a tri-plane structure for capturing geometric features, with an MLP parameterized by $\theta$ for signed distance decoding. Specifically, the tri-plane structure consists of three orthogonal planes (XY, YZ, and XZ) that collectively factorize a dense 3D volume grid. For any query point $q \in \mathbb{R}^3$, the signed distance function $\Phi: \mathbb{R}^3 \to \mathbb{R}$ is computed as follows: 
\begin{equation}
\Phi(q) = \text{MLP}_\theta \left( \bigoplus_{i \in \{xy, xz, yz\}} \text{Interp}_{\gamma_{i}} (\text{Proj}_{\gamma_{i}}(q)) \right),
\label{eq:triplant}
\end{equation}
where $\gamma = \{\gamma_{i} | \gamma_{i} \in \mathbb{R}^{N^2 \times C}, i \in \{xy, xz, yz\}\}$ denotes the three feature planes with $N^2$ spatial resolution and $C$ feature channels, $\text{Proj}(\cdot)$ represents an orthogonal projection of the query point onto each respective plane, $\text{Interp}(\cdot)$ denotes bilinear interpolation that extracts feature vectors from each plane, and $\oplus$ indicates element-wise addition. 

\subsection{Numerical Gradient}
\label{sec:method:numgrad}

While tri-plane representations enhance geometric feature learning and accelerate training, they introduce a locality issue because spatial features are stored in a grid-based structure. Accordingly, each forward pass only considers neighboring features within a small spatial region, limiting the optimization of tri-plane features to adjacent grids. Following previous works~\cite{neuralpull_ma, chen2023gridpull, implicit_li, koneputugodage2024small}, we initially compute the gradient term $\nabla \Phi(\cdot)$ in Eq.~\ref{eq:pull} using analytical gradients. However, in our tri-plane setting, this direct analytical-gradient variant provides a highly local and cell-dependent optimization signal, which can make the optimization sensitive to grid discretization and lead to less stable convergence, especially for sparse unoriented point clouds. The detailed reasons are explained as below:

Given a query point $q = (x, y, z)$ and a plane resolution $N^2$, the feature extraction for the tri-plane encoding from the $XY$ plane can be achieved via bilinear interpolation. We describe the interpolation in the normalized plane coordinate system as follows. First, we scale the query point by the plane resolution: $q_{xy} = (x, y) \cdot N$, and define the coefficients for bilinear interpolation at the four corners of the grid as:
\begin{equation}
    \beta_x = q_{xy,x} - \lfloor q_{xy,x} \rfloor, \quad \beta_y = q_{xy,y} - \lfloor q_{xy,y} \rfloor,
\end{equation}
where $\beta_x$ and $\beta_y$ are the interpolation coefficients along the $x$- and $y$-axes, respectively. The resulting feature vector $\gamma_{xy}(q_{xy})$ from the $XY$ plane is then given by:
\begin{align}
\gamma_{xy}(q_{xy}) = 
& \gamma_{xy}(\lfloor q_{xy,x}\rfloor, \lfloor q_{xy,y} \rfloor) \cdot (1 - \beta_x) \cdot (1 - \beta_y) + \nonumber \\
& \gamma_{xy}(\lceil q_{xy,x}\rceil, \lfloor q_{xy,y} \rfloor) \cdot \beta_x \cdot (1 - \beta_y) + \nonumber \\
& \gamma_{xy}(\lfloor q_{xy,x}\rfloor, \lceil q_{xy,y} \rceil) \cdot (1 - \beta_x) \cdot \beta_y + \nonumber \\
& \gamma_{xy}(\lceil q_{xy,x} \rceil, \lceil q_{xy,y} \rceil) \cdot \beta_x \cdot \beta_y.
\end{align}
After performing bilinear interpolation on each of the three planes ($XY$, $XZ$, and $YZ$), the final feature vector is obtained by aggregating the features from each plane:
\begin{equation}
    \gamma(q) = \gamma_{xy}(q_{xy}) \oplus \gamma_{xz}(q_{xz}) \oplus \gamma_{yz}(q_{yz}),
\end{equation}
where $\oplus$ denotes channel-wise feature concatenation.

The widely-used analytical gradient~[xxx] through this bilinear tri-plane encoding is available, but it is piecewise-defined and highly local. Within each interpolation cell, the derivative is determined only by the grid vertices associated with that cell. Therefore, the supervision signal from one query is back-propagated only to the grid vertices involved in its current interpolation cells. For sparse unoriented point clouds, where the pulling supervision lacks reliable normal directions and is unevenly distributed over the space, this cell-dependent behavior can make the learned field sensitive to grid discretization and local point density.

To address these practical issues for a smoother pulling direction in our tri-plane setting, we employ a central finite-difference estimate for numerical gradient computation. Specifically, for each query point $q = (x, y, z)$, we estimate the gradient by sampling additional points around $q$ at a small offset $\epsilon$ along each axis of the canonical coordinate system. For instance, the $x$-component of the gradient is computed as:
\begin{equation}
\nabla_x \Phi(q) = \frac{\Phi(q + \epsilon_x) - \Phi(q - \epsilon_x)}{2 \epsilon}, 
\label{eq:central_diff_normal}
\end{equation}
where $\epsilon_x = [\epsilon, 0, 0]$ is the perturbation along the $x$-axis; $q + \epsilon_x$ and $q - \epsilon_x$ represent two sample points along the $x$-axis. Using the same approach for $\nabla_y \Phi(q)$ and $\nabla_z \Phi(q)$, we sample a total of six points around $q$ to construct the complete gradient $\nabla \Phi(q) = (\nabla_x \Phi(q), \nabla_y \Phi(q), \nabla_z \Phi(q))$.

In practice, we set $\epsilon$ according to the applied tri-plane resolution, covering a neighborhood comparable to the grid spacing. Compared with the derivative computed within a single interpolation cell, this numerical gradient estimates local SDF variation over neighboring spatial samples and can involve adjacent cells when crossing cell boundaries. As a result, the pulling supervision is less confined to the grid entries activated by the original query point, providing a smoother and more stable optimization signal for tri-plane-based SDF learning.

\subsection{Progressive Tri-Plane Expansion}
\label{sec:method:progressive}

We further propose a coarse-to-fine optimization scheme for surface reconstruction, which progressively expands tri-plane resolution over $n$ stages to prevent the neural SDF's convergence to local minima. Starting with a low-resolution tri-plane facilitates efficient learning of the global shape context and provides a good initialization for subsequent higher-resolution refinements of surface details.

Let the finest tri-plane resolution be $N^2$. We begin the optimization with a tri-plane of resolution $R = \left(\frac{N}{2^n}\right)^2$, where $R$ denotes the current tri-plane resolution, and increase the resolution by a factor of 2 at each stage. We achieve the resolution upsampling via bilinear interpolation. Furthermore, to ensure that the queries for numerical gradients remain within a valid vicinity, we dynamically adjust the perturbation value $\epsilon$ to be inversely correlated with the current resolution: $\epsilon = \frac{1}{2R}$. 
This progressive refinement of the tri-plane resolution helps stabilize training, allowing the network to capture both global and fine-grained surface details while avoiding local minima.

\subsection{Query Locations Sampling}
\label{sec:method:data}

To learn a neural SDF from point clouds, we need to sample query-target point pairs~\cite{neuralpull_ma}. For an extensive sampling of random query positions around each point $p_j$ in $P$, we follow an isotropic Gaussian distribution $\mathcal{N}(p_j, \sigma^2)$ to sample query locations. We randomly sample 25 query points based on this distribution, where $\sigma^2$ controls the sampling range around the surface. The value of $\sigma^2$ is adaptively set as the squared distance between $p_j$ and its 50th nearest neighbor, indicating the local point density around $p_j$.

However, this data sampling strategy is insufficient for the learning of our tri-plane-based hybrid SDF representations. The abovementioned procedure guides the object's surface, while neglecting distant regions. Since our method encodes spatial information on a tri-plane structure, this lack of sufficient guidance can lead to inaccurate or random predictions for the signed distances in distant areas. To address this issue, we introduce another complementary sampling strategy of randomly sampling points in a unit cube $[-1, 1]^3$, aligning with the scale of training data. Accordingly, we not only ensure that every tri-plane entity is trained, but also prevent untrained regions of the tri-plane grid from negatively affecting the learning process.

\subsection{Optimization}
Our proposed method trains a neural SDF by learning to pull a query location $q_i$ towards its nearest neighbor, \emph{i.e.}, the target surface point $t_i$ in the point cloud $P$. Following the computations in Eq.~\ref{eq:pull}, after each training iteration, a query point $q_i$ will be moved to a new position $q'_i$ closer to the point cloud surface, guided by the learned distance value and numerical gradients. Particularly, we train the neural layers of NumGrad-Pull in an end-to-end manner by minimizing the squared Euclidean distances ($\ell_2$ norm) between the pulled query locations and their corresponding nearest neighbor targets. The overall loss is formulated as follows:
\begin{equation}
\mathcal{L} = \frac{1}{I} \sum_{i=1}^{I} \|q'_i - t_i\|^2,
\end{equation}
where $I$ is the number of sampled query points. Similar to prior works~\cite{neuralpull_ma,ma2022reconstructing,jin2023multi} built upon the Neural-Pull paradigm, we do not include the Eikonal~\cite{gropp2020implicit} regularization term, which is typically used to enforce a globally exact signed distance field across the entire space. Instead, our primary objective is to recover the zero-level surface through local pulling dynamics, leveraging the predicted value and gradient direction to pull sampled queries toward nearby surface points. Since the pulling dynamics mainly rely on these local quantities, applying the Eikonal term globally across the sampled space is not strictly necessary for accurate zero-level-set reconstruction.

\section{Experiments}
\label{sec:exp}

We conduct experiments to evaluate the performance of our NumGrad-Pull for surface reconstruction from raw point clouds, presenting results on both synthetic point clouds and real scanned data. We also perform ablation studies to validate the main components of NumGrad-Pull and analyze the impact of various design choices.

\subsection{Implementation Details}
\label{sec:implementation}
We initialize the tri-plane with a resolution of $8^2$ and a feature dimension of 32. The tri-planes are then expanded over three stages at the 3k, 8k, and 12k iterations, ultimately reaching a resolution of $32^2$. To decode the tri-plane features, we use a three-layer fully connected network with a hidden dimension of 128. Additionally, we initialize the network parameters using the geometric network initialization scheme proposed in \cite{gropp2020implicit}, which approximates the signed distance function of a sphere. For optimization, we use the Adam optimizer~\cite{diederik2014adam}, with an initial learning rate of 0.001 for the MLP and 0.05 for the tri-plane parameters. All the experiments are conducted on an Nvidia GeForce RTX 3090 GPU. 

\begin{table}
    \centering
    \caption{Comparisons on ABC~\cite{abc_koch} and FAMOUS~\cite{point2surf_erler} datasets. The metric is chamfer distance ($CD_{\ell_2}$) scaled by $10^3$.}
    \centering
    \begin{adjustbox}{width=\columnwidth}
    \small
    \begin{tabular}{lcccccc}
\toprule
\multirow{2}{*}{Dataset} & NP~\cite{neuralpull_ma}   & PCP~\cite{pcp_ma}  & SIREN~\cite{siren_sitzmann} & DIGS~\cite{ben2022digs} & IF~\cite{implicit_li}   & \multirow{2}{*}{Ours}  \\
&\footnotesize{ICML 2021}&\footnotesize{CVPR 2022}&\footnotesize{NeurIPS 2020}&\footnotesize{CVPR 2022}&\footnotesize{ECCV 2024}&\\
\midrule
ABC     & 0.95 & 2.52 & 0.22  & 0.21 & 0.11 & \textbf{0.09} \\
FAMOUS  & 1.00 & 0.37 & 0.25  & 0.15 & 0.08 & \textbf{0.04} \\
\midrule
Mean    & 0.98 & 1.45 &  0.24 & 0.18 & 0.10 & \textbf{0.07} \\
\bottomrule
\end{tabular}
    \end{adjustbox}
    \label{tab:abc_famous}
\end{table}

\begin{table*}
    \centering
    \caption{Comparisons on the ShapeNet dataset~\cite{shapenet_chang}. The metric is chamfer distance ($CD_{\ell_2}$) scaled by $10^3$.}
    \centering
    \begin{adjustbox}{width=.8\textwidth}
    \small
    \begin{tabular}{l|c|cccccccc|c}
\toprule
Method & Venue & Display & Lamp  & Plane & Cabinet & Vessel & Table & Chair & Sofa  & Avg.  \\ 
\midrule
SPSR~\cite{kazhdan2006poisson} & SGP 2006 & 2.730   & 2.270 & 2.170 & 3.630   & 2.540  & 3.830 & 2.930 & 2.760 & 2.860 \\
NP~\cite{neuralpull_ma} & ICML 2021 & 0.390   & 0.800 & 0.080 & 0.260   & 0.220  & 0.600 & 0.540 & 0.120 & 0.380 \\
LPI~\cite{chen2022latent} & ECCV 2022    & 0.080   & 0.172 & 0.060 & 0.179   & 0.092  & 0.436 & 0.187 & 0.164 & 0.171 \\
PCP~\cite{pcp_ma} & CVPR 2022   & 0.887   & 0.380 & 0.065 & 0.153   & 0.079  & 0.131 & 0.110 & 0.086 & 0.136 \\
GP~\cite{chen2023gridpull} & ICCV 2023    & 0.082   & 0.347 & 0.007 & 0.112   & 0.033  & 0.052 & 0.043 & 0.015 & 0.086 \\
IF~\cite{implicit_li} & ECCV 2024     & \textbf{0.009}   & 0.019 & 0.045 & 0.055   & \textbf{0.005}  & 0.025 & 0.070 & 0.027 & 0.032 \\ 
\midrule
\rowcolor{Gray} Ours & N/A   & 0.024 & \textbf{0.013} & \textbf{0.012} & \textbf{0.032} & 0.012 & \textbf{0.022} & \textbf{0.024} & \textbf{0.020}  & \textbf{0.020}      \\ 
\bottomrule
\end{tabular}
    \end{adjustbox}
    \label{tab:shapenet_cd2}
\end{table*}

\begin{table*}
    \centering
        \caption{Comparison on the Surface Reconstruction Benchmark~\cite{williams2019deep}.}
    \centering
    \begin{adjustbox}{width=\textwidth}
    \begin{tabular}{l|cccc|cccc|cccc|cccc|cccc}
\toprule
\multirow{2}{*}{Method} & \multicolumn{4}{c|}{Anchor}                                           & \multicolumn{4}{c|}{Daratech}                                         & \multicolumn{4}{c|}{DC}                                               & \multicolumn{4}{c|}{Gargoyle}                                         & \multicolumn{4}{c}{Lord Quas}                                         \\
                        & $CD_{\ell_1}$ & $HD$  & $d_{\overrightarrow{C}}$ & $d_{\overrightarrow{H}}$ & $CD_{\ell_1}$ & $HD$  & $d_{\overrightarrow{C}}$ & $d_{\overrightarrow{H}}$ & $CD_{\ell_1}$ & $HD$  & $d_{\overrightarrow{C}}$ & $d_{\overrightarrow{H}}$ & $CD_{\ell_1}$ & $HD$  & $d_{\overrightarrow{C}}$ & $d_{\overrightarrow{H}}$ & $CD_{\ell_1}$ & $HD$  & $d_{\overrightarrow{C}}$ & $d_{\overrightarrow{H}}$ 
 \\ 
\cline{1-21} 
SPSR~\cite{kazhdan2006poisson}                    & 0.60      & 14.89 & 0.60                     & 14.89                  & 0.44      & 7.24  & 0.44                     & 7.24                   & 0.27      & 3.10  & 0.27                     & 3.10                   & 0.26      & 6.80  & 0.26                     & 6.80                   & 0.20      & 4.61  & 0.20                     & 4.61                   \\
IGR~\cite{gropp2020implicit}                     & 0.22      & 4.71  & 0.12                     & 1.32                   & 0.25      & 4.01  & 0.08                     & 1.59                   & 0.17      & 2.22  & 0.09                     & 2.61                   & 0.16      & 3.52  & 0.06                     & 0.81                   & 0.12      & 1.17  & 0.07                     & 0.98                   \\
SIREN~\cite{siren_sitzmann}                   & 0.32      & 8.19  & 0.10                     & 2.43                   & 0.21      & 4.30  & 0.09                     & 1.77                   & 0.15      & 2.18  & 0.06                     & 2.76                   & 0.17      & 4.64  & 0.08                     & 0.91                   & 0.17      & 0.82  & 0.12                     & 0.76                   \\
VisCo~\cite{pumarola2022visco}                   & 0.21      & 3.00  & 0.15                     & 1.07                   & 0.21      & 4.06  & 0.14                     & 1.76                   & 0.15      & 2.22  & 0.09                     & 2.76                   & 0.17      & 4.40  & 0.11                     & 0.96                   & 0.12      & 1.06  & 0.07                     & 0.64                   \\
SAP~\cite{peng2021shape}                     & 0.12      & 2.38  & 0.08                     & 0.83                   & 0.26      & 0.87  & 0.04                     & 0.41                   & 0.07      & 1.17  & 0.04                     & 0.53                   & 0.07      & 1.49  & 0.05                     & 0.78                   & 0.05      & 0.98  & 0.04                     & 0.51                   \\
NP~\cite{neuralpull_ma}                     & 0.122     & 3.243 & 0.061                    & 3.208                  & 0.375     & 3.127 & 0.746                    & 3.267                  & 0.157     & 3.541 & 0.242                    & 3.523                  & 0.080     & 1.376 & 0.063                    & 0.475                  & 0.064     & 0.822 & 0.053                    & 0.508                  \\
GP~\cite{chen2023gridpull}                      & 0.093     & 1.804 & 0.066                    & 0.460                  & 0.062     & \textbf{0.648} & 0.039                    & 0.293                  & 0.066     & 1.103 & 0.036                    & 0.539                  & 0.063     & 1.129 & 0.045                    & 0.700                  & 0.047     & 0.569 & 0.031                    & 0.370                  \\
DIGS~\cite{ben2022digs}                    & 0.063     & 1.447 & 0.030                    & 0.270                  & \textbf{0.049}     & 0.858 & 0.025                    & 0.441                  & 0.042     & 0.667 & 0.022                    & 0.729                  & 0.047     & 0.971 & 0.028                    & 0.271                  & 0.031     & 0.496 & 0.017                    & 0.181                  \\
NSH~\cite{NSH} & 0.085 & \textbf{0.521} & \textbf{0.018} & 0.117 & 0.134 & 0.765 & \textbf{0.014} & 0.245 & 0.091 & 0.528 & \textbf{0.018} & 0.288 & 0.082 & 0.527 & 0.023 & 0.140 &
  0.117 & 0.617 & \textbf{0.013} & \textbf{0.053} \\
IF~\cite{implicit_li}                      & 0.052     & 1.232 & 0.025                    & 0.265                  & 0.051     & 0.751 & 0.028                    & 0.423                  & 0.041     & 0.815 & 0.019                    & 0.724                  & 0.044     & 1.089 & \textbf{0.022}                    & 0.246                  & \textbf{0.030}     & 0.554 & 0.014                    & 0.230                  \\ 
\midrule

\rowcolor{Gray} Ours  & \textbf{0.051}  & 1.194 & 0.041  & \textbf{0.113} & 0.050 & 0.730 & 0.036 & \textbf{0.236} & \textbf{0.026} & \textbf{0.215} & 0.032 & \textbf{0.286} & \textbf{0.033} & \textbf{0.416} & 0.040 & \textbf{0.127} & 0.036 & \textbf{0.293} & 0.039 & 0.127 \\ 
\bottomrule
\end{tabular}
    \end{adjustbox}
    \label{tab:srb}
\end{table*}

\begin{figure*}
    \centering
    \includegraphics[width=\textwidth]{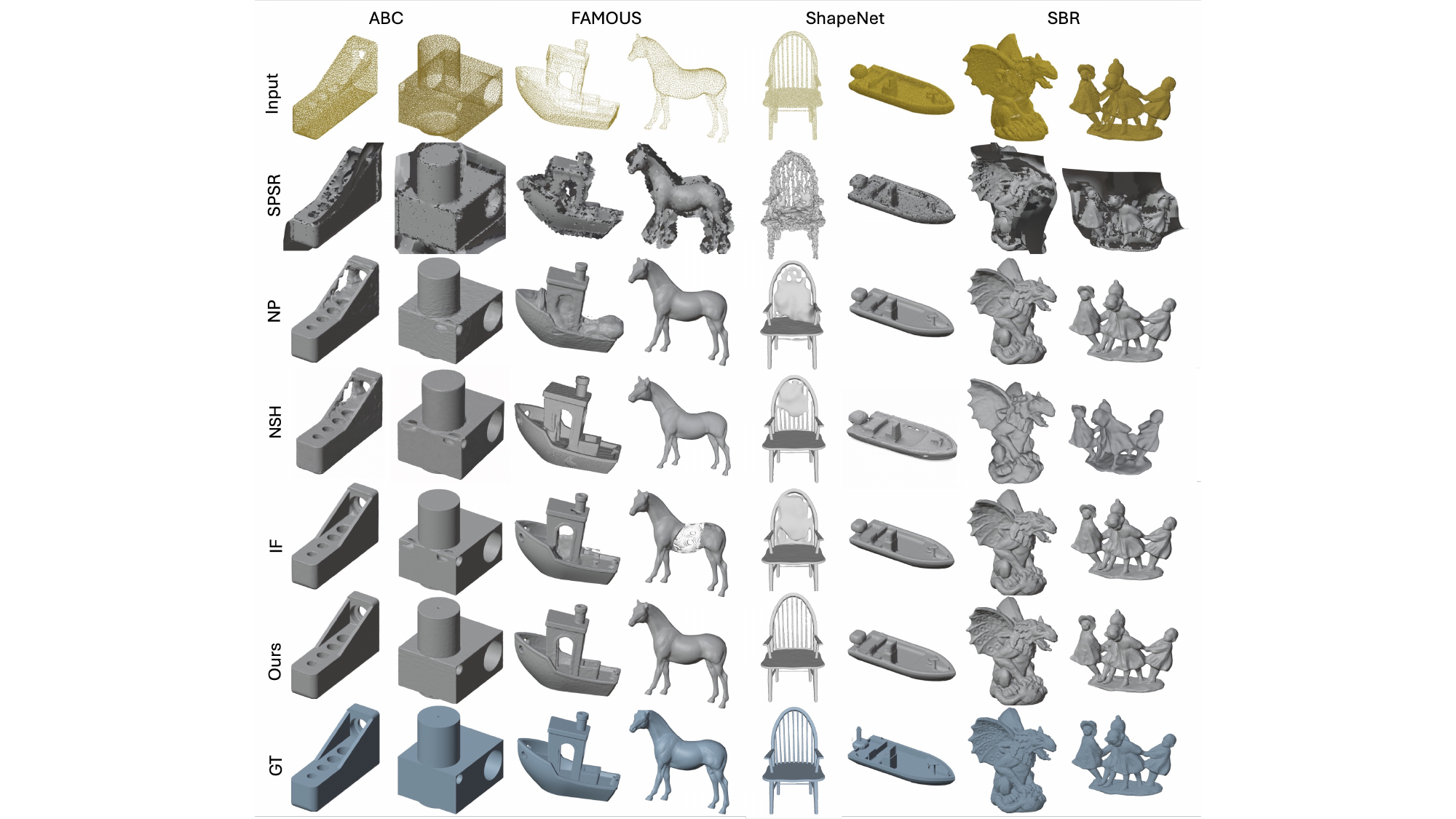}
    \vspace{2mm}
    \caption{Visual comparisons of surface reconstruction quality on the synthetic datasets (ABC~\cite{abc_koch}, FAMOUS~\cite{point2surf_erler}, ShapeNet~\cite{shapenet_chang}) and the real-world scan dataset (SRB~\cite{williams2019deep}).}
    \label{fig:compare}
    \vspace{1mm}
\end{figure*}

\begin{figure*}[t]
    \centering
    \includegraphics[width=0.95\textwidth]{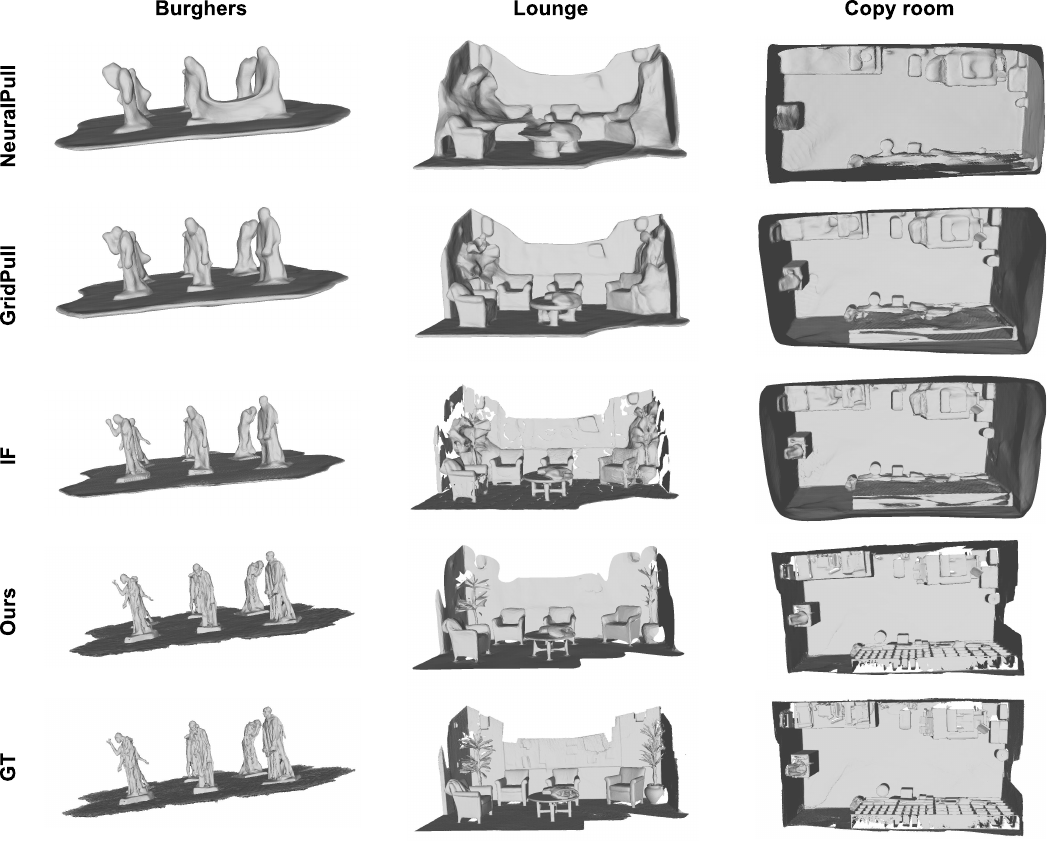}
    \caption{Visual comparisons of surface reconstruction quality on the 3D Scene dataset~\cite{10.1145/2461912.2461919}.}
    \label{fig:wild_compare}
\end{figure*}

\subsection{Surface Reconstruction for Shapes}

\subsubsection{Datasets and Metrics}

For surface reconstruction of general shapes from raw point clouds, we follow the evaluation protocol used in prior work~\cite{neuralpull_ma,implicit_li} and assess our method on three widely used datasets: ShapeNet~\cite{shapenet_chang}, ABC~\cite{abc_koch}, and FAMOUS~\cite{point2surf_erler}. For ShapeNet, we use the test split defined in \cite{liu2019learning} and the pre-processed data provided in \cite{mescheder2019occupancy}. Following the footsteps of~\cite{point2surf_erler}, we employ the same train/test split and pre-processed data for the ABC and FAMOUS datasets.

For performance evaluation, we follow the approach of previous works~\cite{neuralpull_ma,implicit_li}, sampling $1\times 10^5$ points from both the reconstructed surfaces and the ground truth meshes for the ShapeNet dataset and $1 \times 10^4$ points for the ABC and FAMOUS datasets. We use Chamfer distance (CD), measured by $\ell_2$ norm to quantify the distance between our reconstructions and the ground truth surfaces.

\subsubsection{Comparison}

We compare the performance of our proposed NumGrad-Pull with several state-of-the-art methods, including SPSR~\cite{kazhdan2006poisson}, PCP~\cite{pcp_ma}, LPI~\cite{chen2022latent}, SIREN~\cite{siren_sitzmann}, Neural-Pull~\cite{neuralpull_ma}, DIGS~\cite{ben2022digs}, Grid-Pull~\cite{chen2023gridpull},
NSH~\cite{NSH} and the recent IF~\cite{implicit_li}. Results on the ABC and FAMOUS datasets are presented in Tab.~\ref{tab:abc_famous}, highlighting the superiority of our method: NumGrad-Pull outperforms the latest state-of-the-art by margins of 0.02 and 0.04 on the ABC and FAMOUS datasets, respectively. Furthermore, our method shows strong performance in Tab.~\ref{tab:shapenet_cd2} on the ShapeNet dataset, which is a challenging and comprehensive benchmark comprising over 3,000 testing objects. Our NumGrad-Pull achieves the best average surface reconstruction quality across all methods, outperforming the state-of-the-art IF~\cite{implicit_li} method by 0.012 and achieving the best results in six out of eight categories. These quantitative results demonstrate the effectiveness of our approach in improving surface reconstruction quality.

In Fig.~\ref{fig:compare}, we present qualitative comparisons with other approaches~\cite{kazhdan2006poisson,neuralpull_ma,implicit_li}, showcasing that our method effectively reconstructs object surfaces from unoriented point clouds across varying levels of sparsity and noise. Notably, our method reliably captures detailed structures, as seen in the chair sample from the ShapeNet dataset. Similarly, our approach produces intact surfaces without the breaks observed in other methods, as illustrated in the second example from the ABC dataset. These results further highlight the effectiveness and robustness of our approach.

\subsection{Surface Reconstruction for Real Scans}
\subsubsection{Datasets and Metrics}
For surface reconstruction from real point cloud scans, we follow the evaluation protocol of VisCo~\cite{pumarola2022visco} on the Surface Reconstruction Benchmark (SRB)~\cite{williams2019deep}, which includes five challenging real-world scans. We adopt Chamfer and Hausdorff distances ($CD_{\ell_1}$ and $HD$) between the reconstruction meshes and the ground truth as the quality metrics. Following the state-of-the-art method~\cite{implicit_li}, we also report the corresponding uni-directional distances ($d_{\overrightarrow{C}}$ and $d_{\overrightarrow{H}}$) between the reconstructed meshes and the input noisy point clouds to evaluate the performance in preserving scanned geometries.

\subsubsection{Comparison}
We compare our method with various prior methods using the specified evaluation metrics, with per-object results presented in Tab.~\ref{tab:srb}.  Our method performs well across multiple metrics and objects, particularly achieving superior reconstruction quality on the Anchor and the Gargoyle object. To further substantiate the results, we provide visualizations of the Gargoyle (7th column) and DC (8th column) samples in Fig.~\ref{fig:compare}. Our method captures finer details, such as the rings on the Gargoyle’s wings, and preserves accurate geometry even in areas with corruptions in the original scan, demonstrating robust detail retention and reconstruction fidelity.

\subsection{Qualitative Results on 3D Scenes}
\label{sec:wild_vis}
We further evaluate surface reconstruction on the real-world 3D Scene dataset~\cite{10.1145/2461912.2461919}. Fig.~\ref{fig:wild_compare} presents qualitative comparisons on three representative scenes. Compared with NeuralPull~\cite{neuralpull_ma}, GridPull~\cite{chen2023gridpull}, and IF~\cite{implicit_li}, our approach consistently produces higher-quality surfaces. Moreover, fine structures like thin parts and sharp boundaries are better preserved, leading to clearer separation among nearby instances in crowded scenes.

This improvement can be attributed to our tri-plane design. Prior MLP-only SDFs are primarily supervised using nearest-point target pairs, where the supervision signal is highly local and lacks global constraints, particularly for large, scene-level scans. In our NumGrad-Pull method, we leverage dense near-surface Gaussian samples to learn accurate local surface details, and additionally exploits uniformly sampled points within the unit cube to provide global free-space supervision for tri-plane learning. As a result, our tri-plane design not only encodes local geometry from dense near-surface supervision, but also leverages uniform samples for global regularization, leading to more stable, robust, and generalizable reconstruction performance on real-world 3D scenes.
From the scene-level perspective, this hybrid representation is more computationally efficient and practically scalable than using deep MLPs: spatial information is explicitly encoded by the factorized planes, while the decoder remains lightweight. Such explicit-implicit representations have been widely adopted for efficient neural field and scene modelling~\cite{muller2022instant,li2023neuralangelo}, while our tri-plane-based design can also effectively support scene-level reconstruction from larger-scale 3D data.

\subsection{Ablation Studies}

\begin{figure}
    \centering
    \includegraphics[width=\linewidth]{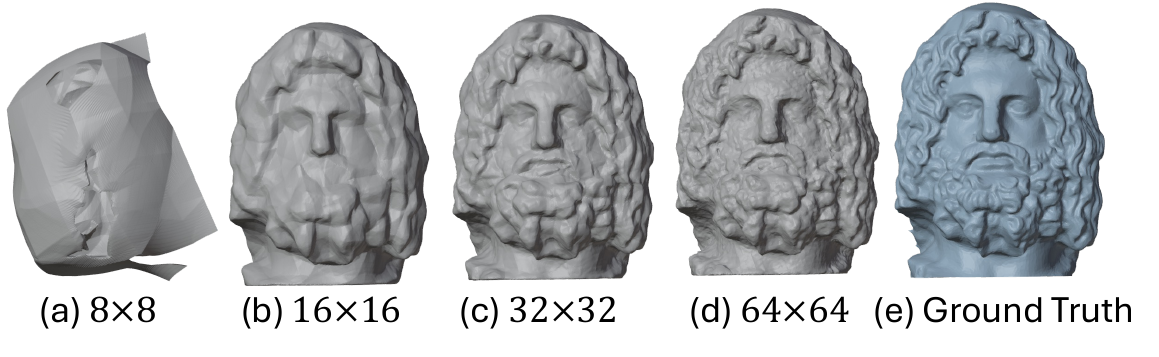}
    \caption{Visual results of surface reconstruction using different tri-plane resolutions. Low-resolution (8 and 16) tri-planes tend to over-smooth geometric details, while high-resolution tri-planes can preserve sharper structures but may also introduce unstable high-frequency surface variations (64) when regularization is insufficient (related quantitative results can be found in Table~\ref{tab:abl:resolution} and Table~\ref{tab:reg_tradeoff}).}
    \label{fig:abl_reso}
\end{figure}

\begin{table}
    \centering
    \caption{Effect of different tri-plane configurations on surface reconstruction for the FAMOUS dataset~\cite{point2surf_erler} compared with recent methods, measured by $CD_{\ell_2}$ and scaled by $10^4$.}
    \begin{adjustbox}{width=\columnwidth}
    \small
    \begin{tabular}{lcccccc}
\toprule
Metric & 8 & 16  & 32 & 64 & NP~\cite{neuralpull_ma} & IF~\cite{implicit_li}\\
\midrule
$CD_{\ell_2}$ $\downarrow$ & 9.31 & 0.45 & 0.39  & \textbf{0.35} & 1.00 & 0.80 \\
Speed (iter/s) $\uparrow$ & 219 & \textbf{231} & 227 & 225 & 122 & 42 \\ 
\bottomrule
\end{tabular}
    \end{adjustbox}
    \label{tab:abl:resolution}
\end{table}

\subsubsection{Effect of Tri-plane Resolution}
Considering that tri-plane resolution impacts surface reconstruction quality, we empirically evaluate the performance of NumGrad-Pull using various resolutions, 8, 16, 32, and 64, on the FAMOUS dataset. Visual results are shown in Fig.~\ref{fig:abl_reso}, and quantitative results are provided in Tab.~\ref{tab:abl:resolution}. Our findings indicate that as tri-plane resolution increases, surface reconstruction quality (measured by $CD_{\ell_2}$) improves. The most substantial gain is observed when increasing the resolution from 8 to 16, while increasing to 32 and 64 resolution yields marginal improvements. The significant quality boost from 8 to 16 suggests that an $8 \times 8$ tri-plane lacks sufficient capacity to store the necessary information for accurate surface reconstruction, as shown in Fig.~\ref{fig:abl_reso}~(b). Higher resolutions enhance the network's ability to capture fine structures; however, with a $64 \times 64$ tri-plane, while the metric show slight improvements, the visual results reveal unwanted surface unevenness, indicating that excessive high-frequency noise is being captured. Overall, to balance reconstruction quality with robustness against noise, we apply a $32 \times 32$ tri-plane for our main experiments.

We also present efficiency comparisons between recent methods with our NumGrad-Pull using different tri-plane resolutions in Tab.~\ref{tab:abl:resolution}. Since our proposed tri-plane-based neural SDF can significantly improve query speed, we achieve a 1.8× speedup over Neural-Pull~\cite{neuralpull_ma} and a 5.4× speedup over IF~\cite{implicit_li}, while retaining superior surface reconstruction quality. Furthermore, the tri-plane representation exhibits excellent scalability with respect to resolution: as the tri-plane resolution increases, the query speed remains stable. This stability is attributed to the feature interpolation process being an $O(1)$ operation, ensuring that query speed is unaffected by resolution changes.
\begin{table}
      \centering
      \caption{Reconstruction quality ($CD_{\ell_1} \times 10^3 \downarrow$) under different tri-plane resolutions and regularization values ($\lambda$) at 50,000 iterations.}
      \label{tab:reg_tradeoff}
      \begin{tabular}{lcccc}
          \toprule
          Resolution & $\lambda=0$ & $\lambda=10^{-4}$ & $\lambda=5\times10^{-4}$ & $\lambda=10^{-3}$ \\
          \midrule
          16  & \textbf{38.42} & 38.55 & 39.12 & 40.28 \\
          32  & 35.18 & \textbf{34.82} & 35.07 & 36.41 \\
          64  & 35.26 & 33.54 & \textbf{33.12} & 34.05 \\
          128 & 36.88 & 33.42 & \textbf{32.85} & 33.18 \\
          \bottomrule
      \end{tabular}
  \end{table}
We further analyze the relations between tri-plane resolutions and associated regularization values in Table~\ref{tab:reg_tradeoff}. The results indicate that higher resolutions (64 or 128) are beneficial only under appropriate regularization. For example, Res-128 improves from 36.88 to 32.85 in $CD_{\ell_1}\times10^3$ when $\lambda$ is increased from 0 to $5\times10^{-4}$. Complementing Fig.~\ref{fig:abl_reso}, this table further reveals that, although higher resolutions can yield better CD scores, they are also highly sensitive to the choice of regularization strength. In practice, we select the tri-plane resolution as a balanced configuration that jointly considers reconstruction accuracy, visual fidelity, and robustness to hyperparameter variations.

\subsubsection{Effect of Query Ratio}
\label{sec:abl:hybrid_eff}
To further analyze the efficiency trade-off of our tri-plane formulation, we evaluate different values of the \emph{query ratio}, ranging from $1/2$ to $1/32$, which determine the fraction of query points used to train the tri-plane-based SDF representation in Eq.~\ref{eq:triplant}.

\begin{table}
    \centering
    \caption{Effect of different query point ratios used to train the Tri-plane-based SDF representation.}
    \begin{adjustbox}{width=\columnwidth}
    \small
    
\begin{tabular}{lcccc}
\toprule
Ratio & ABC~\cite{abc_koch} & FAMOUS--3DBenchy~\cite{point2surf_erler} & ShapeNet--Sofa~\cite{shapenet_chang} & SRB--Gargoyle~\cite{williams2019deep}\\
\midrule
1/2  & 0.11 & 0.08 & 0.024 & 0.044 \\
1/4  & 0.11 & 0.05 & 0.021 & 0.041 \\
1/8  & \textbf{0.09} & \textbf{0.04} & \textbf{0.020} & \textbf{0.033} \\
1/16 & 0.12 & \textbf{0.04} & 0.022 & 0.037 \\
1/32 & 0.14 & 0.09 & 0.029 & 0.041 \\
\bottomrule
\end{tabular}

    \end{adjustbox}
    \label{tab:abl:hybrid_ratio}
\end{table}

\begin{table}
    \centering
    \caption{Effect of each proposed module of our method on surface reconstruction for the FAMOUS dataset, measured by $CD_{\ell_2}$, scaled by $10^4$.}
    \begin{adjustbox}{width=\columnwidth}
    \small
    \begin{tabular}{lccccc}
\toprule
Model    & w/o Data & w/o Tri-plane & w/o NumGrad & w/o Progressive. & $CD_{L2}$ \\
\midrule
baseline & \ding{55} & \ding{55} & \ding{55} & \ding{55} & 11.35 \\
A & \ding{55} & \ding{51} & \ding{51} & \ding{51} &  50.21 \\ 
B & \ding{51} & \ding{55} & \ding{51} & \ding{51} &  0.41 \\ 
C & \ding{51} & \ding{51} & \ding{55} & \ding{51} & 331.84 \\ 
D & \ding{51} & \ding{51} & \ding{51} & \ding{55} & 0.45 \\ 
Full & \ding{51} & \ding{51} & \ding{51} & \ding{51} & \textbf{0.39} \\       
\bottomrule
\end{tabular}
    \end{adjustbox}
    \label{tab:abl:component}
\end{table}

We report the quantitative results on five representative datasets in Tab.~\ref{tab:abl:hybrid_ratio}. Specifically, the ratio of $1/8$ achieves the best overall performance across all evaluated cases. When the ratio is too large (\emph{e.g.}, $1/2$ or $1/4$), the model tends to rely excessively on dense tri-plane queries, which increases sensitivity to high-frequency noise and does not consistently improve reconstruction quality. Conversely, when the ratio is too small (\emph{e.g.}, $1/16$ or $1/32$), the supervision and feature representation of the tri-plane branch become insufficient, leading to degraded reconstruction quality, as evidenced by the noticeable performance drop on ABC and SRB samples. The results suggest that a moderate query ratio can effectively balance the benefits of fast, robust, and generalizable tri-plane-based SDF modeling. We thus adopt the $1/8$ ratio as the default setting in our main experiments.

\subsubsection{Effect of Each Module}
\label{sec:modules}
\begin{figure}
      \centering
      \begin{adjustbox}{width=\columnwidth}
          \includegraphics{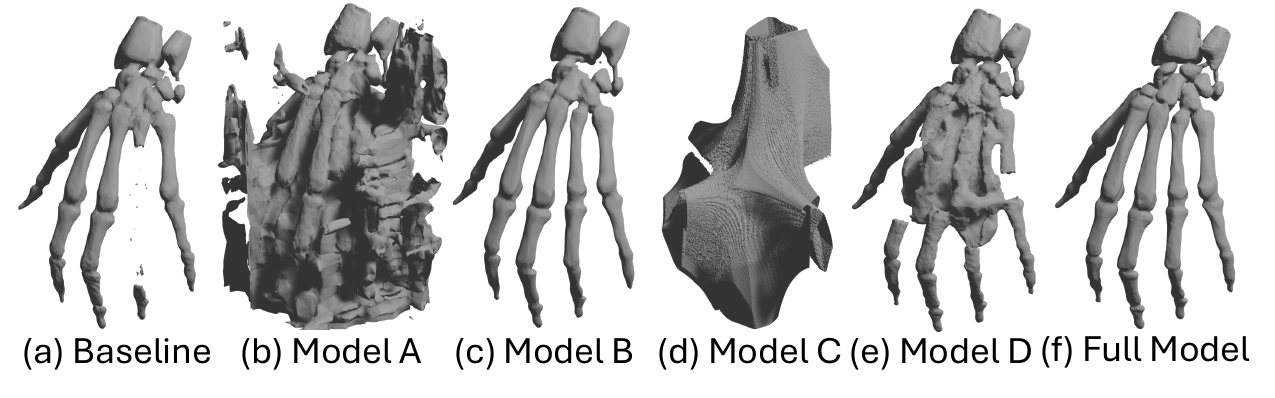}
      \end{adjustbox}
      \caption{Visual results of using different model variants.}
      \label{fig:abl_components}
  \end{figure}

We conduct a detailed ablation study to validate the effectiveness of each proposed module, including the following model variants: A) full model without our data sampling strategy; B) full model without the tri-plane structure; C) full model without numerical gradients; and D) full model without progressive tri-plane expansion. These variants are compared with a baseline model Neural-Pull~\cite{neuralpull_ma} and our complete model incorporating all proposed modules. We evaluate the performances on the FAMOUS dataset, provide quantitative results in Tab.~\ref{tab:abl:component}, and present a visualization of the \textit{hand} sample in Fig.~\ref{fig:abl_components}.
Our full model achieves the best visual and quantitative results, capturing all essential details without artifacts. In contrast, each variant shows varying degrees of performance degradation. \textbf{Model~A} lacks a comprehensive data sampling strategy, sampling query points only near the object surface, similar to Neural-Pull. This model suffers from insufficient supervision for regions far from the surface, leading to poorly trained tri-plane features at distant positions and causing random predictions and artifacts. \textbf{Model~B} replaces the tri-plane with an 8-layer fully connected network, achieving results close to our full model but with reduced reconstruction detail and training stability. While it retains more structural detail than the baseline by employing numerical gradients, it is approximately 3.2 times slower due to the slower query speed of the fully connected network compared to the tri-plane structure. \textbf{Model~C} fails to train effectively due to the adoption of analytical gradients, which only propagate supervision to local grids and causes instability and corrupted outputs. \textbf{Model~D} struggles without using our proposed progressive expansion strategy, which is crucial for preventing the network from getting stuck in local optima. This experimental comparison highlights the necessity of our strategy for the effective training of the tri-plane structure. Overall, these results demonstrate that each of the proposed modules is essential, contributing to the superior performance of the full model.

\begin{figure}[ht]
    \centering
    \begin{adjustbox}{width=0.85\columnwidth}
    \includegraphics{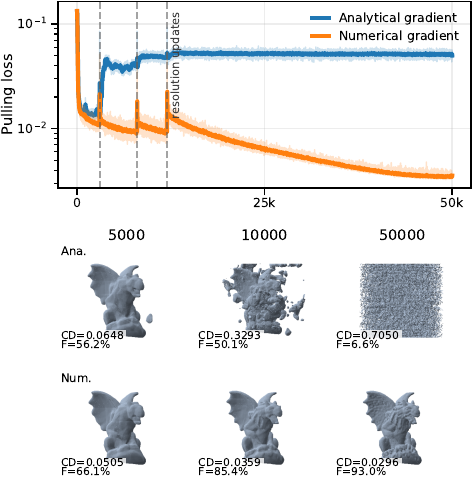}
    \end{adjustbox}
    \caption{Comparison between the analytical-gradient and numerical-gradient variants on the SRB-Gargoyle sample. Top: pulling-loss curves. Bottom: intermediate reconstruction results at iterations 5000, 10000, and 50000. Each reconstruction panel reports the F-Score@0.005 and CD $\times 10$, evaluated under the same Chamfer-L1 protocol as in Tab.~\ref{tab:srb}.}
    \label{fig:gradient_variant}
\end{figure}

To further explain why \textbf{Model C} fails, Fig.~\ref{fig:gradient_variant} examines the optimization behavior of different gradient variants on the SRB-Gargoyle sample. The results indicate that numerical gradients are better aligned with our progressive tri-plane representation. After each resolution update of the tri-plane, analytical gradients derived from interpolated features become unstable, causing the loss to rise and the reconstruction quality to degrade. In contrast, numerical gradients consistently provide stable local geometric directions, enabling smooth convergence throughout training. At iteration 50000, the analytical-gradient variant fails with CD $=0.7050$ and F-Score@0.005 $=6.6\%$, while the numerical-gradient variant achieves CD $=0.0296$ and F-Score@0.005 $=93.0\%$. These results clearly demonstrate the synergy between numerical gradients and the progressive tri-plane design.

\subsubsection{Feature Representation Ablation}
To further evaluate the role of the tri-plane representation, we replace it with two alternative feature encoders: a parameter-matched dense 3D grid and an InstantNGP-style multi-resolution hash grid encoder. All variants are trained for 5,000 iterations with comparable feature-encoder parameter counts.

\begin{table}[ht]
  \centering
  \caption{Effect of different feature representations. CD is reported as CD $\times 10^3$.}
  \label{tab:feature_repr_ablation}
  \begin{adjustbox}{width=\columnwidth}
  \begin{tabular}{lccccc}
  \toprule
  Representation & Params & CD $\downarrow$ & F-Score $\uparrow$ & Loss $\downarrow$ & Memory \\
  \midrule
  Dense 3D Grid & 257K & 7.15 & 64.46 & 0.0121 & \textbf{159 MiB} \\
  Hash Grid & 300K & 18.40 & 48.97 & 0.0098 & 185 MiB \\
  Tri-plane (Ours) & 259K & \textbf{6.90} & \textbf{66.87} & \textbf{0.0084} & 160 MiB \\
  \bottomrule
  \end{tabular}
  \end{adjustbox}
\end{table}

\begin{figure}[ht]
    \centering
    \includegraphics[width=\linewidth]{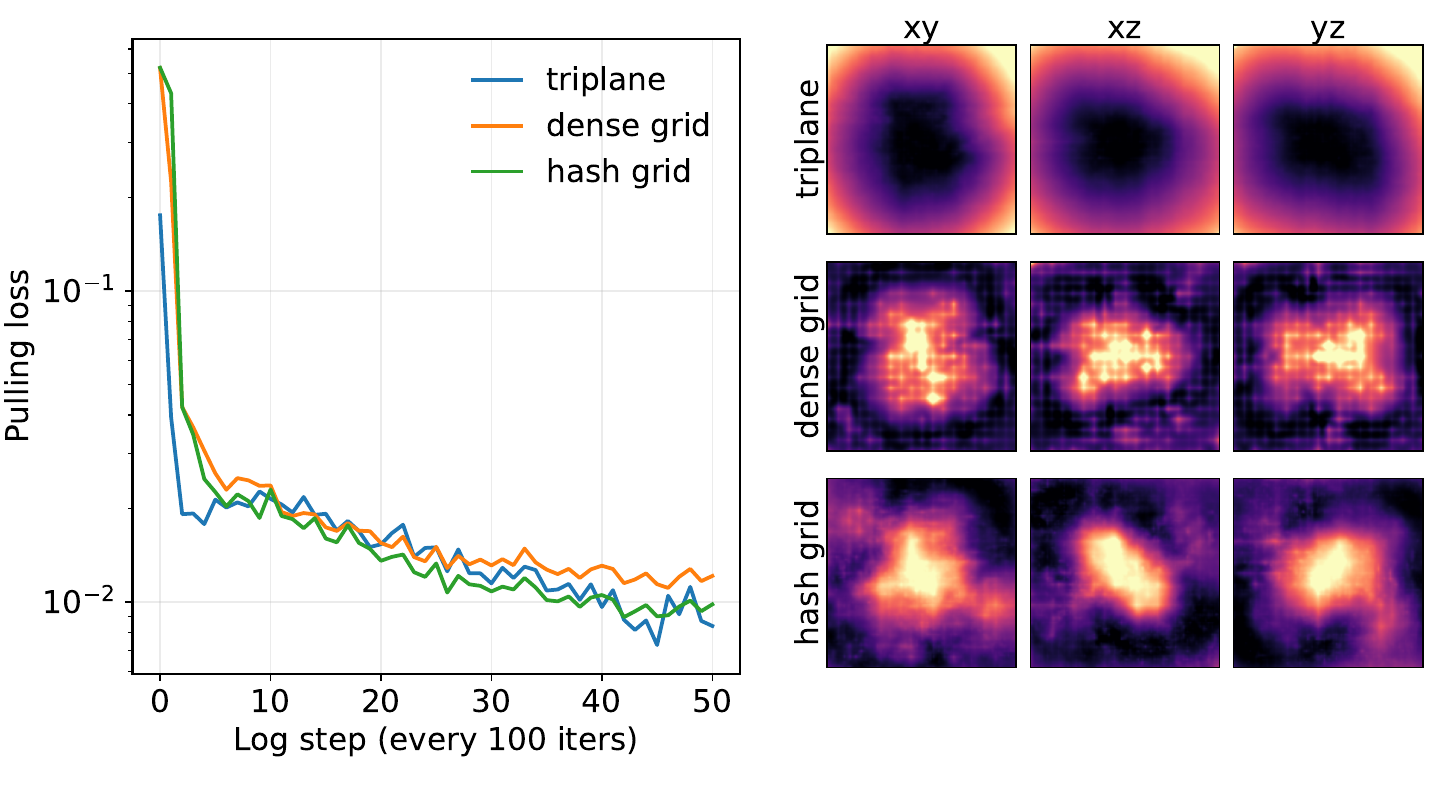}
    \includegraphics[width=\linewidth]{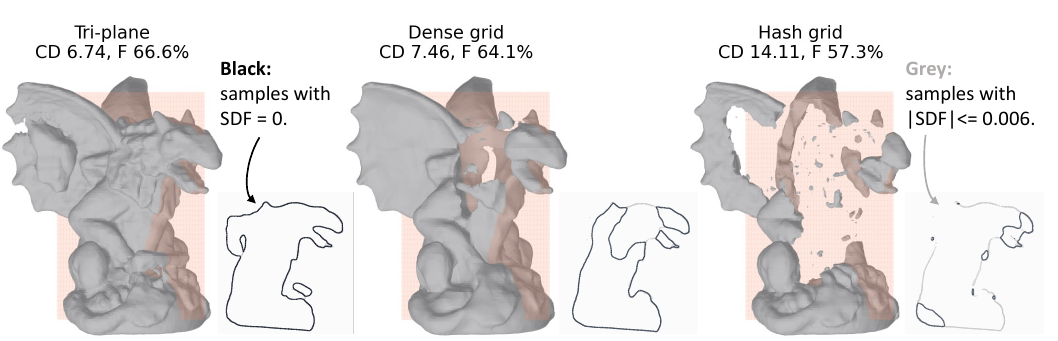}
    \caption{Visualizations of training-loss curves, feature-slices, reconstructed meshes, and zero-level-set slices (slicing planes in pink) for different feature representations. Smoother and more coherent feature slices indicate more stable spatial feature learning under the pulling objective.}
    \label{fig:feature-slice}
\end{figure}

As shown in Tab.~\ref{tab:feature_repr_ablation}, the tri-plane representation achieves the best overall reconstruction quality among the tested variants. As the dense 3D grid's memory grows cubically with resolution, the parameter-matched version has to compromise at a relatively low spatial resolution with worse reconstruction quality. The hash-grid encoder is compact, but its learned feature distribution is less spatially smooth under the pulling objective, leading to worse CD and F-Score in this setting. The training-loss curves and feature-slice visualizations in Fig.~\ref{fig:feature-slice} further show that the tri-plane learns smoother, more coherent spatial feature distributions. These results support our choice of tri-planes as a good trade-off between accuracy, memory efficiency, and stable feature learning.

\subsubsection{Computational Costs}
\label{sec:compute_tradeoffs}
We further analyze the computational costs of analytical and numerical gradient computation. Since the gradient estimator affects training but not mesh extraction, we compare peak GPU memory and iteration speed under the same Gargoyle setting with 1,000 training iterations and batch size 5,000. We report Neural-Pull~\cite{neuralpull_ma} and IF~\cite{implicit_li} with their default analytical-gradient settings as reference baselines, and directly compare the analytical- and numerical-gradient variants of our NumGrad-Pull.

\begin{table}[t]
  \centering
  \caption{Computational costs of analytical- and numerical-gradient variants.}
  \label{tab:compute_main}
   \begin{adjustbox}{width=\columnwidth}
    \small
   \begin{tabular}{llccc}
    \toprule
     Method & Gradient & Memory (MiB) $\downarrow$ & iter/s $\uparrow$\\
      \midrule
      Neural-Pull & Analytical & 270 & 232.1 \\
      IF          & Analytical & 950 & 51.8 \\
      NumGrad-Pull (Ours) & Analytical & 160 & 242.1 \\
      NumGrad-Pull (Ours) & Numerical & \textbf{157} & \textbf{302.3} \\
      \bottomrule
  \end{tabular}
    \end{adjustbox}
  \end{table}

As shown in Tab.~\ref{tab:compute_main}, NumGrad-Pull maintains low memory usage under both gradient settings due to the applied lightweight tri-plane design. More importantly, the numerical-gradient variant does not introduce a serious practical disadvantage in our implementation: the six offset SDF queries are computed jointly in a single batched forward pass, thereby avoiding the costly autograd chain that the analytical-gradient variant must construct through the input coordinates and tri-plane interpolation. As a result, NumGrad-Pull with numerical gradients achieves the best memory usage and iteration speed among the comparisons in Tab.~\ref{tab:compute_main}.

\subsubsection{Step Size Sensitivity Analysis}
\label{sec:step_size_sensitivity}
\begin{figure}[ht]
    \centering
    \includegraphics[width=0.7\linewidth]{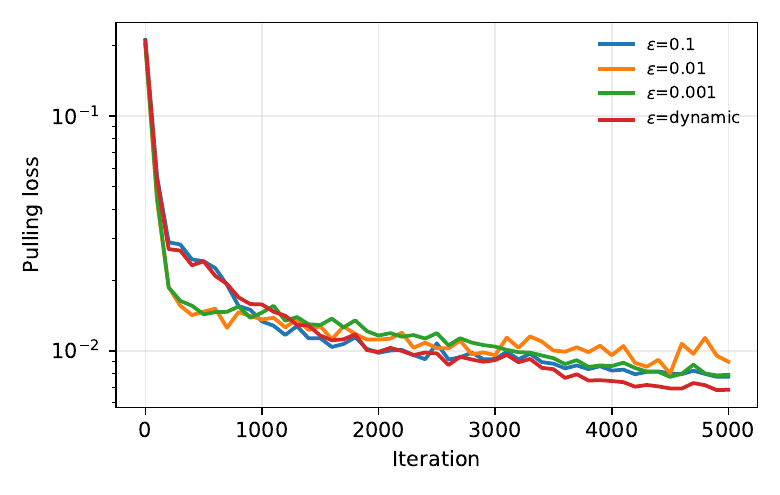}
    \caption{Loss curves of using different numerical-gradient step sizes. The dynamic step size $\epsilon = 1/(2R)$ achieves stable convergence and the lowest
  final pulling loss compared with fixed step sizes.}
    \label{fig:step_size_loss}
\end{figure}

We further analyze the sensitivity of the numerical-gradient step size by comparing three fixed step sizes, $\epsilon \in \{0.1, 0.01, 0.001\}$, with our dynamic strategy $\epsilon = 1/(2R)$, where $R$ denotes the current tri-plane resolution.
As shown in Tab.~\ref{tab:step_size_ablation}, the dynamic step size gives the best overall reconstruction accuracy and trained objective among the tested settings. The fixed step sizes are reasonably stable, but their performances vary with the chosen scale. A large step size may over-smooth local geometric variations, while an overly small step size can make the finite-difference estimate too local and less robust. Notably, reducing $\epsilon$ to very small values leads to performance degradation, as the numerical gradient increasingly approximates the analytical gradient. In contrast, our applied dynamic strategy adapts the finite-difference interval to the feature-grid resolution, providing a more consistent balance between smoothing and locality. The corresponding loss curves in Fig.~\ref{fig:step_size_loss} further show that the dynamic strategy maintains stable convergence and achieves the lowest final pulling loss.

\begin{table}[ht]
  \centering
  \caption{Sensitivity analysis of the numerical-gradient step size. CD is reported as CD $\times 10^3$.}
  \label{tab:step_size_ablation}
  \begin{tabular}{lcccc}
  \toprule
  Step Size $\epsilon$ & CD $\downarrow$ & F-Score $\uparrow$ & Pulling Loss $\downarrow$ & iter/s \\
  \midrule
  $0.1$   & 6.07 & 56.70 & 0.0078 & 116.2 \\
  $0.01$  & 6.00 & 55.69 & 0.0090 & 116.6 \\
  $0.001$ & 6.30 & 52.70 & 0.0079 & 116.9 \\
  Dynamic $1/(2R)$ & \textbf{5.59} & \textbf{57.71} & \textbf{0.0068} &
  \textbf{116.9} \\
  \midrule
  $0.0001$ & 6.92 & 51.28 & 0.0093 & 116.8 \\
  Analytical gradient & 7.05 & 50.62 & 0.0095 & 116.6\\ 
  \bottomrule
  \end{tabular}
\end{table}

\subsubsection{Pulling Strategy Comparison}
We study our numerical-gradient pulling strategy and existing alternatives, continuous gradient-based pulling~\cite{neuralpull_ma}, multi-step pulling~\cite{noda2024multipull}, gradient-corrected pulling~\cite{implicit_li}, and hash-grid-based acceleration~\cite{zhou2024fast}, on the reconstruction of the SRB Gargoyle sample. We evaluate pulling accuracy via the final endpoint error (``Pull.~Err.'', \emph{i.e.}, the mean Euclidean distance from each pulled query point to its corresponding target surface point), convergence speed via iterations per second (``iter/s''), and reconstruction quality via CD and F-Score.
All variants share an identical backbone and training budget, differing only in the pulling scheme. 
As shown in Table~\ref{tab:pulling}, our method achieves the best reconstruction quality (CD 2.95, F-Score 93.1\%), pulling accuracy (0.0035), and high efficiency (303.3~iter/s). The gradient-corrected method achieves comparable quality but at much higher memory cost and $5\times$ slower throughput. Although the hash-grid method is fast and memory-efficient, its larger pulling error reflects a less stable pulling direction and consequently degraded reconstruction quality. In contrast, our tri-plane representation offers a more suitable spatial structure for reliable cross-cell numerical gradient estimation.

\begin{table}[ht]
  \centering
  \caption{Pulling Strategy Comparison. CD is reported as CD $\times 10^3$.}
  \label{tab:pulling}
  \begin{adjustbox}{width=\columnwidth}
  \begin{tabular}{lrrrrr}
  \toprule
  Pulling Strategy Variant & CD $\downarrow$ & F-Score $\uparrow$ & Pull. Err. $\downarrow$ & iter/s $\uparrow$ & Mem. (MiB) $\downarrow$ \\
  \midrule
  Continuous gradient-based~\cite{neuralpull_ma} & 8.69 & 55.7 & 0.0127 & 254.1 & 270.3 \\
  Multi-step~\cite{noda2024multipull} & 3.63 & 89.0 & 0.0040 & 181.6 & 159.9 \\
  Gradient-corrected~\cite{implicit_li}  & 3.32 & 88.1 & 0.0047 & 64.6 & 789.2 \\
  Hash-grid-based acceleration~\cite{zhou2024fast} & 10.35 & 48.6 & 0.0220 & \textbf{309.2} & \textbf{61.9} \\
  Ours (numerical gradient) & \textbf{2.95} & \textbf{93.1} & \textbf{0.0035} & 303.3 & 159.7 \\
  \bottomrule
  \end{tabular}
  \end{adjustbox}
\end{table}

\begin{figure}[ht]
    \centering
\includegraphics[width=0.9\linewidth]{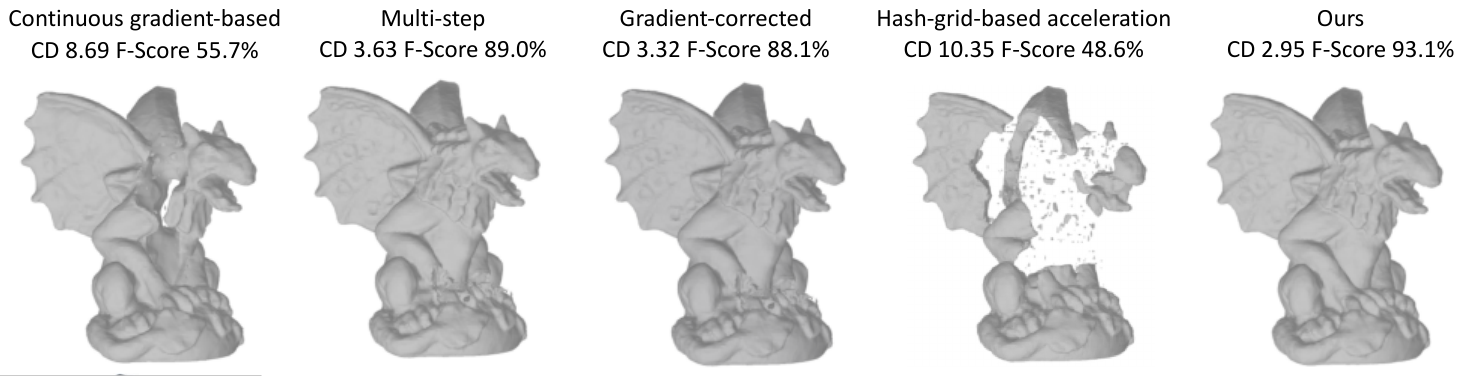}
    \caption{Reconstruction results of different pulling strategies.}
    \label{fig:pull}
\end{figure}

\subsubsection{Robustness Analysis}
\label{sec:robustness}
We further evaluate robustness under Gaussian noise and sparse-input settings. For noise, we perturb the input point cloud case with Gaussian noise from $0\%$ to $1.0\%$ of the bounding-box diagonal. For sparsity, we vary the number of input points using nested random subsets from 4K to 20K, under a fixed $0.5\%$ Gaussian perturbation.

\begin{figure}[ht]
  \centering
  \includegraphics[width=\linewidth]{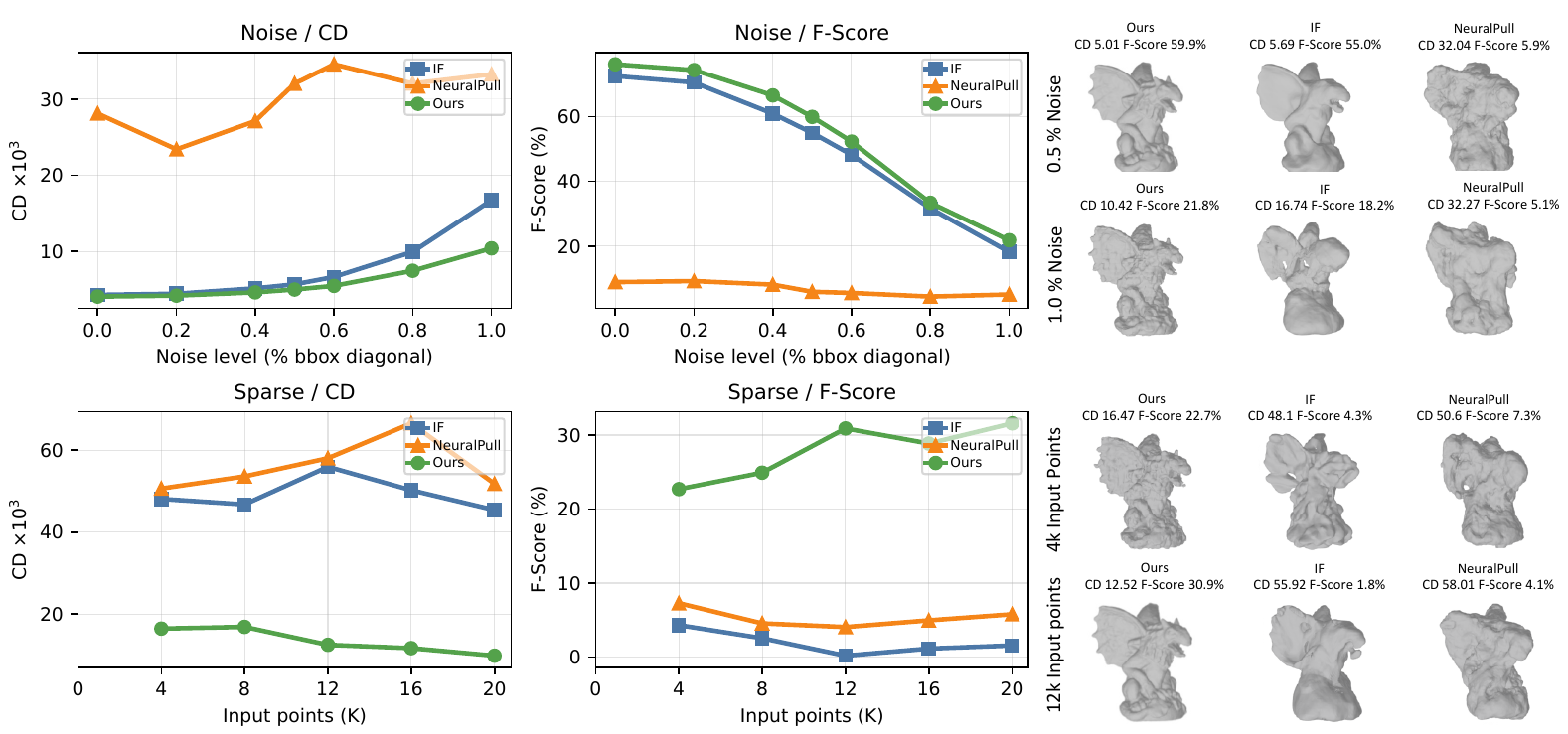}
  \caption{Robustness analysis under noisy and sparse inputs.  The top row shows CD, F-Score curves, and reconstruction results, under increasing Gaussian noise; and the bottom row shows CD, F-Score curves, and reconstruction results, under varying sparse-input settings.}
  \label{fig:robustness}
\end{figure}

As shown in Fig.~\ref{fig:robustness}, NumGrad-Pull remains stable under increasing Gaussian noise and outperforms the baselines across the tested noise levels. At $1.0\%$ noise, NumGrad-Pull obtains CD $=10.42\times10^{-3}$ and F-Score $=21.84\%$, compared with CD $=16.74\times10^{-3}$ and F-Score $=18.21\%$ for IF. For sparse inputs, NumGrad-Pull also maintains better reconstruction quality across the tested input-point levels. At the 20K input-point setting, NumGrad-Pull achieves CD $=9.87\times10^{-3}$ and F-Score $=31.58\%$, compared with CD $=45.38\times10^{-3}$ and F-Score $=1.57\%$ for IF.

\section{Conclusion}
\label{sec:conclusion}

In this paper, we introduce NumGrad-Pull, a novel approach for high-fidelity surface reconstruction from unoriented point clouds. Our method employs a hybrid explicit–implicit tri-plane representation to improve query speed and enhance reconstruction fidelity. Furthermore, we propose a numerical gradients-based technique to stabilize the training process and a progressive expansion strategy to fully leverage the capabilities of the tri-plane structure. Extensive experiments demonstrate that our NumGrad-Pull consistently outperforms state-of-the-art methods across various challenging scenarios, showcasing significant effectiveness in surface reconstruction problems. We believe NumGrad-Pull has strong potential to improve 3D content creation and streamline the conversion of raw 3D data into production-ready formats.  While we do not foresee any immediate harmful uses, we emphasize the importance of users exercising caution to minimize potential risks, given that surface reconstruction tools could also be leveraged for malicious purposes.

\bibliographystyle{IEEEtran}
\bibliography{ref}

\newpage

\vfill

\end{document}